\documentclass[10pt,twocolumn,letterpaper]{article}

\usepackage{cvpr}
\usepackage{times}
\usepackage{epsfig}
\usepackage{graphicx}
\usepackage{amsmath}
\usepackage{amssymb}
\usepackage{algorithm}
\usepackage{algorithmic}
\usepackage{multirow}

\usepackage[breaklinks=true,bookmarks=false]{hyperref}

\cvprfinalcopy 

\def\cvprPaperID{6629}

\begin{document}

\title{ChamNet: Towards Efficient Network Design through Platform-Aware Model Adaptation}

\author{Xiaoliang Dai$^{1}$, 
Peizhao Zhang$^{2}$, 
Bichen Wu$^{3}$,
Hongxu Yin$^{1}$, 
Fei Sun$^{2}$, 
Yanghan Wang$^{2}$, 
Marat Dukhan$^{2}$,\\
Yunqing Hu$^{2}$,
Yiming Wu$^{2}$,
Yangqing Jia$^{2}$,
Peter Vajda$^{2}$, 
Matt Uyttendaele$^{2}$,
Niraj K. Jha$^{1}$\\
1: Princeton University, Princeton, NJ \\
2: Facebook Inc., Menlo Park, CA \\
3: University of California, Berkeley, Berkeley, CA
}
\maketitle

\begin{abstract}
This paper proposes an efficient neural network (NN) architecture design methodology called 
Chameleon that honors given resource constraints. 
Instead of developing new building blocks or using computationally-intensive reinforcement learning algorithms, 
our approach leverages existing efficient network building blocks and focuses on exploiting hardware traits and adapting computation resources to fit target latency and/or energy constraints. 
We formulate platform-aware NN architecture search in an optimization framework and propose a novel 
algorithm to search for optimal architectures aided by efficient accuracy and resource (latency and/or 
energy) predictors. At the core of our algorithm lies an accuracy predictor built atop Gaussian 
Process with Bayesian optimization for iterative sampling.  
With a one-time building cost for the predictors, our algorithm produces state-of-the-art model 
architectures on different platforms under given constraints in just minutes. 
Our results show that adapting computation resources to building blocks is critical to model 
performance. Without the addition of any bells and whistles, our models achieve significant accuracy 
improvements against state-of-the-art hand-crafted and automatically designed architectures.
We achieve 73.8\% and 75.3\% top-1 accuracy on ImageNet at 20ms latency on a mobile CPU and
DSP. At reduced latency, our models achieve up to 8.5\% (4.8\%) and 6.6\% (9.3\%) absolute top-1 
accuracy improvements compared to MobileNetV2 and MnasNet, respectively,  on a mobile CPU (DSP), 
and 2.7\% (4.6\%) and 5.6\% (2.6\%) accuracy gains over ResNet-101 and ResNet-152,
respectively, on an Nvidia GPU (Intel CPU).
\end{abstract}

\section{Introduction}
Neural networks (NNs) have led to state-of-the-art performance in myriad 
areas, such as computer vision, speech recognition, and machine translation. 
Due to the presence of millions of parameters and floating-point operations (FLOPs), NNs are 
typically too computationally-intensive to be deployed on resource-constrained 
platforms.  Many efforts have been made recently to design compact NN 
architectures.  Examples include NNs presented in \cite{mobilenetv2, shufflenetv2, mnasnet}, 
which have significantly cut down the computation cost and achieved a much more 
favorable trade-off between accuracy and efficiency.
However, a compact model design still faces challenges upon deployment in 
real-world applications \cite{netadapt}:
\begin{itemize}
    \itemsep0em
    \item 
    Different platforms have diverging hardware characteristics. It is hard 
for a single NN architecture to run optimally on all the different platforms.  
For example, a Hexagon v62 DSP prefers convolution operators with a channel 
size that is a multiple of 32, as shown later, whereas this may not be
the case on another platform.
    \item  
    Real-world applications may face very different constraints.  For example, 
real-time video frame analysis may have a very strict latency constraint, 
whereas Internet-of-Things (IoT) edge device designers may care more about 
run-time energy consumption for longer battery life.  It is infeasible to have 
one NN that can meet all these constraints simultaneously.  This makes 
it necessary to adapt the NN architecture to the specific use scenarios before 
deployment. 
\end{itemize}

There are two common practices for tackling these challenges. The first 
practice is to manually craft the architectures based on the characteristics 
of a given platform. However, such a trial-and-error methodology might be too 
time-consuming for large-scale cross-platform NN deployment and may not 
be able to effectively explore the design space.  Moreover, it also requires 
substantial knowledge of the hardware details and driver libraries.  The other 
practice focuses on platform-aware neural architecture search (NAS) and 
sequential model-based optimization (SMBO) \cite{nasnet}. Both NAS and SMBO 
require computationally-expensive network training and measurement of network 
performance metrics (e.g., latency and energy) throughout the entire 
search and optimization process. For example, the latency-driven mobile NAS 
(MNAS) architecture requires hundreds of GPU hours to develop~\cite{mnasnet}, 
which becomes unaffordable when targeting numerous platforms with various 
resource budgets. Moreover, it may be difficult to implement the new 
cell-level structures discovered by NAS because of their 
complexity~\cite{nasnet}.

In this paper, we propose an efficient, scalable, and automated NN 
architecture adaptation methodology. We refer to this methodology as Chameleon. 
It does not rely on new cell-level building blocks nor does it use 
computationally-intensive reinforcement learning (RL) techniques.
Instead, it takes into account the traits of the hardware platform to
allocate computation resources accordingly when searching the design space for 
the given NN architecture with existing building blocks.  This adaptation reduces search time.
It employs predictive models (namely accuracy, latency, and energy predictors) 
to speed up the entire search process by enabling immediate performance metric 
estimation. The accuracy and energy predictors incorporate Gaussian process 
(GP) regressors augmented with Bayesian optimization and imbalanced quasi 
Monte-Carlo (QMC) sampling. It also includes an operator latency look-up table 
(LUT) in the latency predictor for fast, yet accurate, latency
estimation. It consistently delivers higher accuracy and less run-time latency against 
state-of-the-art handcrafted and automatically searched models across 
several hardware platforms (e.g., mobile CPU, DSP, Intel CPU, and Nvidia GPU) 
under different resource constraints.  For example, on a Samsung Galaxy S8 with Snapdragon 835 CPU, our adapted model yields 1.7\% higher top-1 accuracy and 1.75$\times$ speedup compared to MnasNet~\cite{mnasnet}.  

\begin{figure*}[t]
\begin{center}
\includegraphics[width=133mm]{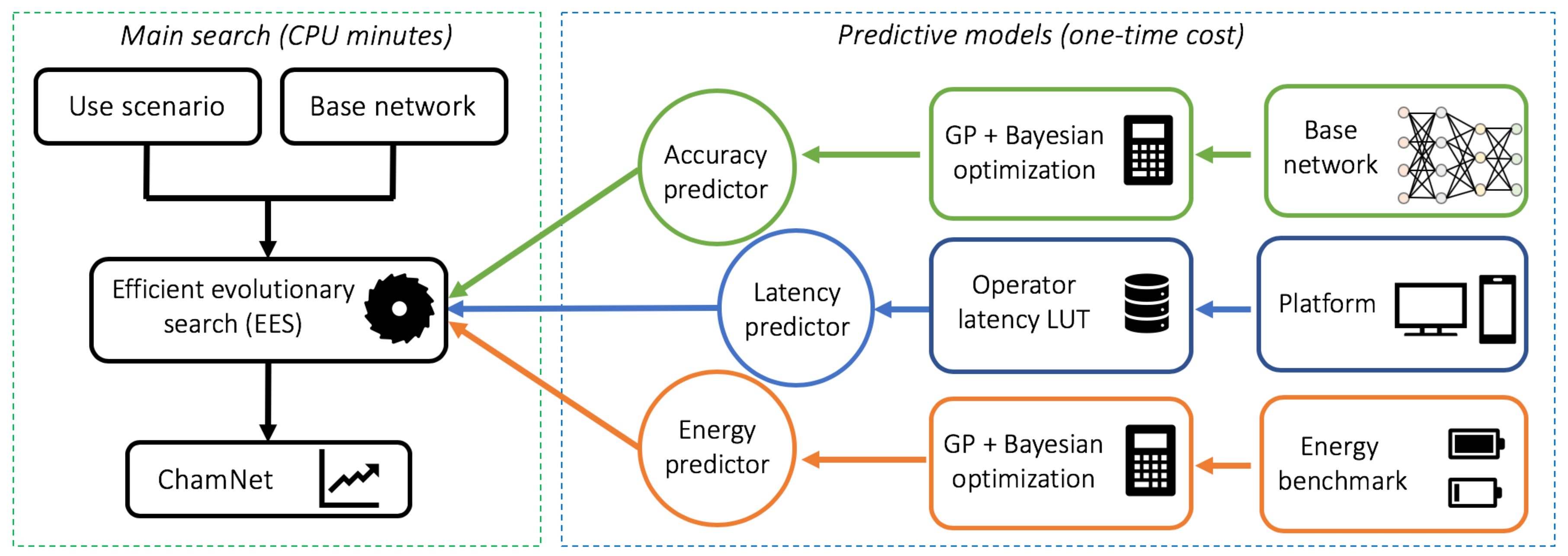}
\caption{An illustration of the Chameleon adaptation framework}
\label{fig:framework}
\end{center}
\vspace{-1mm}
\end{figure*}

Our contributions can be summarized as follows:

\begin{enumerate}\itemsep-0em

\item We show that computation distribution is critical to model performance. By 
leveraging existing efficient building blocks, we adapt models with significant improvements over 
state-of-the-art hand-crafted and automatically searched models under a wide spectrum of devices 
and resource budgets. 

\item We propose a novel algorithm that searches for optimal architectures through efficient
accuracy and resource predictors. At the core of our algorithm lies an accuracy predictor built
based on GP with Bayesian optimization that enables a more effective search over a
similar space to RL-based NAS. 

\item Our proposed algorithm is efficient and scalable. With a one-time building cost, 
it only takes minutes to search for models under different platforms/constraints, thus 
making them suitable for large-scale heterogeneous deployment.

\end{enumerate}

\section{Related work}
Efficient NN design and deployment is a vibrant field.  We summarize the related work next.

\textbf{Model simplication:} An important direction for efficient NN design 
is model simplification.  Network pruning~\cite{NetPrune, structured_sparsity, 
NeST, netprune2, energy_aware, lstmprune} has been a popular approach 
for removing redundancy in NNs.  For example, NetAdapt~\cite{netadapt} utilizes 
a hardware-aware filter pruning algorithm and achieves up to 1.2$\times$ 
speedup for MobileNetV2 on the ImageNet dataset~\cite{imagenet}. 
AMC~\cite{amc} employs RL for automated model compression and achieves 
1.53$\times$ speedup for MobileNet\_v1 on a Titan XP CPU.  
Quantization~\cite{deepcompression, binary} has also emerged as a powerful tool 
for significantly cutting down computation cost with no or little accuracy 
loss.  For example, Zhu et al.~\cite{tenary} show that there is only 
a 2\% top-5 accuracy loss for ResNet-18 when using a 3-bit representation 
for weights compared to its full-precision counterpart.

\textbf{Compact architecture:} Apart from simplifying existing models, 
handcrafting more efficient building blocks and operators for mobile-friendly 
architectures can also substantially improve the accuracy-efficiency 
trade-offs~\cite{squeezenet, shift}.  For example, at the same accuracy level, 
MobileNet~\cite{mobilenet} and ShuffleNet~\cite{shufflenet} cut down the 
computation cost substantially compared to ResNet~\cite{resnet} 
by utilizing depth-wise convolution and low-cost group convolution, 
respectively.  Their successors, MobileNetV2~\cite{mobilenetv2} and 
ShuffleNetV2~\cite{shufflenetv2}, further shrink the model size while 
maintaining or even improving accuracy.  In order to deploy these models on 
different real-world platforms, Andrew et al.~propose linear scaling 
in~\cite{mobilenet}. This is a simple but widely-used method to accommodate 
various latency constraints. It relies on thinning a network uniformly at 
each layer or reducing the input image resolution.

\textbf{NAS and SMBO:} Platform-aware NAS and SMBO have emerged as a promising 
direction for automating the synthesis flow of a model based on direct
metrics, making it more suitable for deployment~\cite{smbo1, hardware, 
archsearch, progressive, darts, smboalgo}. For example, MnasNet~\cite{mnasnet} yields an 
absolute 2\% top-1 accuracy gain compared to MobileNetV2 1.0x with only 
a 1.3\% latency overhead on Google Pixel 1 using TensorFlow 
Lite.  As for SMBO, Stamoulis et al.~use a Bayesian optimization approach and 
reduce the energy consumed for VGG-19~\cite{vgg} on a mobile device 
by up to 6$\times$.  Unfortunately, it is difficult to scale NAS and SMBO 
for large-scale platform deployment, since the entire search and
optimization needs to be conducted once per network per platform per use case.

\section{Methodology}
We first give a high-level overview of the Chameleon 
framework, after which we zoom into predictive models.

\subsection{Platform-aware Model Adaptation}
We illustrate the Chameleon approach in Fig.~\ref{fig:framework}.  The 
adaptation step takes a default NN architecture and a specific use scenario 
(i.e., platform and resource budget) as inputs and generates an adapted 
architecture as output.  Chameleon searches for a variant of the base NN 
architecture that fits the use scenario through efficient evolutionary search 
(EES).  EES is based on an adaptive genetic algorithm~\cite{adaptiveGA}, where 
the gene of an NN architecture is represented by a vector of hyperparameters 
(e.g., \#Filters and \#Bottlenecks), denoted as $\textbf{x} \in 
\textbf{R}^{n}$, where $n$ is the number of hyperparameters of interest.  In each iteration, EES evaluates the fitness of each 
NN architecture candidate based on inputs from the predictive models, and 
then selects architectures with the highest fitness to breed the next 
generation using mutation and crossover operators.  EES terminates after 
a pre-defined number of iterations.  Finally, Chameleon rests at an adapted NN 
architecture for the target platform and use scenario.

We formulate EES as a constrained optimization problem.  The objective is to 
maximize accuracy under a given resource constraint on a target platform:
\begin{equation}
\textrm{maximize}\ \textit{A}(\textbf{x})
\end{equation}
\begin{equation}
\textrm{subject\ to}\ \textit{F}(\textbf{x}, \textit {plat}) \leq \textit{thres}
\end{equation}
where $A$, $F$, $plat$, and $thres$ refer to mapping of $\textbf{x}$ to 
network accuracy, mapping of $\textbf{x}$ to the network performance 
metric (e.g., latency or energy), target platform, and resource constraint 
determined by the use scenario (e.g., 20ms), respectively. We merge the 
resource constraint as a regularization term in the fitness function $R$
as follows:
\begin{equation}
R = A(\textbf{x}) - [\alpha H(F(\textbf{x}, plat) - thres)]^{w}
\end{equation}
where $H$ is the Heaviside step function, and $\alpha$ and $w$ are positive 
constants.  Consequently, the aim is to find the network gene $\textbf{x}$ 
that maximizes R:
\begin{equation}
    \textbf{x} = \underset{\textbf{x}}{argmax}(R)
\end{equation}
We next estimate $F$ and $A$ to solve the optimization problem.  Choices 
of $F$ depends on constraints of interest. In 
this work, we mainly study $F$ based on direct latency and energy 
measurements, as opposed to an indirect proxy, such as FLOPs, that has been 
shown to be sub-optimal~\cite{netadapt}. Thus, for each NN candidate 
$\textbf{x}$, we need three performance metrics to calculate its 
$R(\textbf{x})$: accuracy, latency, and energy consumption.  

Extracting the above metrics through network training and direct measurements 
on hardware, however, is too time-consuming~\cite{progressive}.  To speed up 
this process, we bypass the training and measurement process by leveraging 
accuracy, latency, and energy predictors, as shown in Fig.~\ref{fig:framework}. 
These predictors enable metric estimation in less than one CPU second. We give details of our accuracy, latency, and energy predictors next.

\subsection{Efficient Accuracy Predictor}
To significantly speed up NN architecture candidate evaluation, we utilize 
an accuracy predictor to estimate the final accuracy of a model without 
actually training it.  There are two desired objectives of such a predictor:
\begin{enumerate}
\itemsep-0em 
\item Reliable prediction: The predictor should minimize the distance between 
predicted and real accuracy, and rank models in the same order as their real 
accuracy.
\item Sample efficiency: The predictor should be built with as few trained 
network architectures as possible. This saves computational resources in
training instance generation.
\end{enumerate}
Next, we explain how we tackle these two objectives through GP regression 
and Bayesian optimization based sample architecture selection for training.

\subsubsection{Gaussian Process Model}
We choose a GP regressor as our accuracy predictor to model $A$ as:
\begin{equation}
\begin{aligned}
A(\textbf{x}_{i}) = & f(\textbf{x}_{i}) + \epsilon_{i},\ i = 1, 2, ..., s \\
 f & (\cdot) \sim \mathcal{G}\mathcal{P}(\cdot | 0, K)\\ 
& \epsilon_{i}  \sim \mathcal{N}(\cdot | 0, \sigma^{2}) \\
\end{aligned}
\end{equation}
where \textit{i} denotes the index of a training vector among \textit{s} training vectors and
$\epsilon_{i}$'s refer to noise variables with independent $\mathcal{N}(\cdot | 0, \sigma^{2})$ 
distributions.  
$f(\cdot)$ is drawn from a GP prior characterized by covariance matrix $K$. We use a radial basis 
function kernel for $K$:
\begin{equation}
K(\textbf{x}, \textbf{x}') = \rm{exp}(-\gamma||\textbf{x} - \textbf{x}'||^{2})
\end{equation}

\begin{figure}
\begin{center}
\includegraphics[width=75mm]{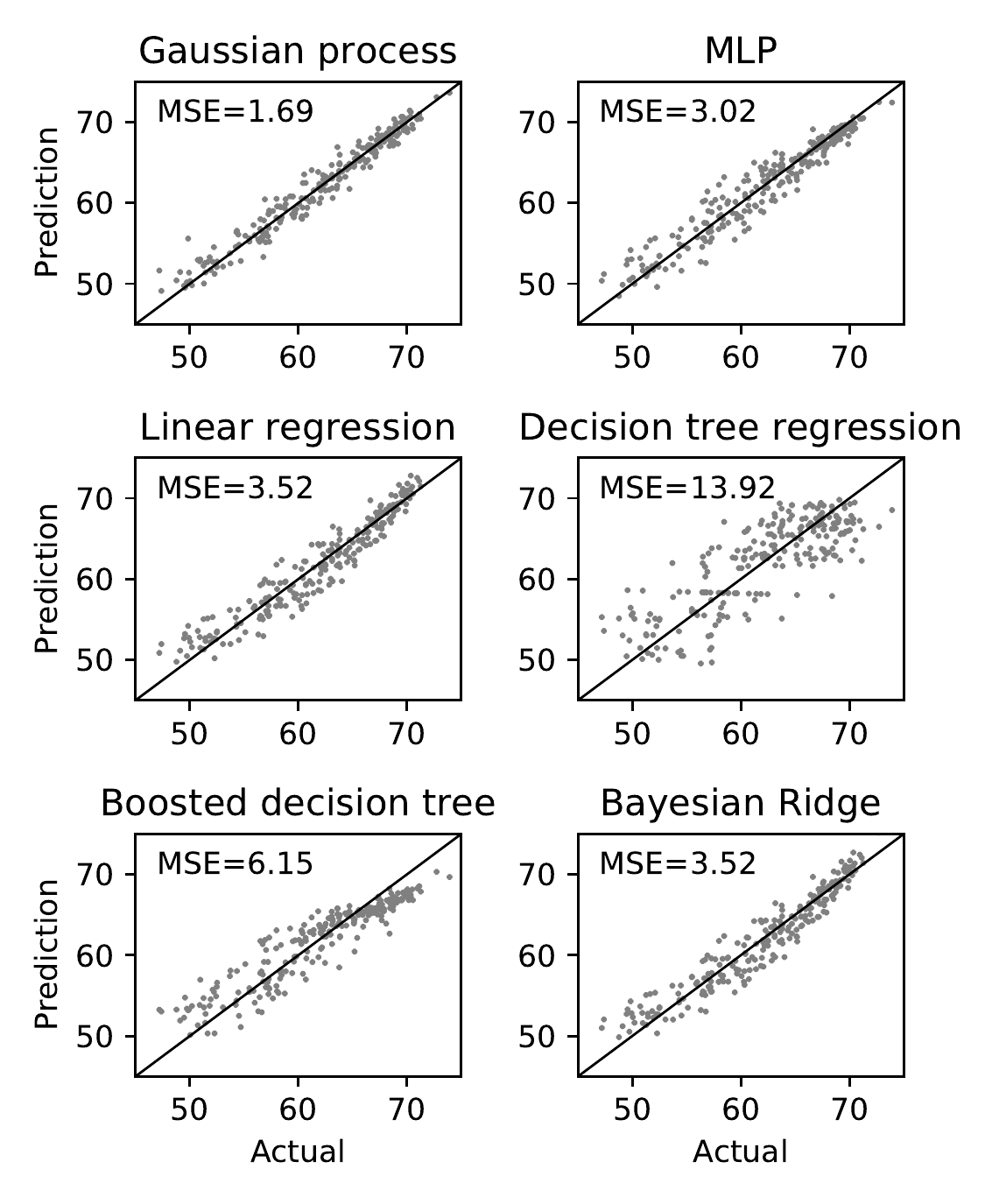}
\caption{Performance comparison of different accuracy prediction models,
built with 240 pretrained models under different configurations.  MSE refers 
to leave-one-out mean squared error.}
\label{fig:accu_pred}
\end{center}
\vspace{-2mm}
\end{figure}

A GP regressor provides two benefits.  First, it offers 
reliable predictions when training data are scarce.  As an example, we compare 
several regression models for MobileNetV2 accuracy prediction in 
Fig.~\ref{fig:accu_pred}.  The GP regressor has the lowest mean squared error 
(MSE) among all six regression models. Second, a GP regressor produces 
predictions with uncertainty estimations, which offers additional guidance for 
new sample architecture selection for training.  This helps boost the 
convergence speed and improves sample efficiency, as shown next.

\begin{algorithm}[h]
   \caption{Steps for building an accuracy predictor}
   \small
   \label{alg:accu_pred}
\begin{algorithmic}
   \STATE {\bfseries Input:} $k$: sample architecture pool size, 
$p$: exploration sample count, $q$: exploitation sample count, $e$: MSE 
threshold
   \STATE Pool = Get $k$ QMC samples from the adaptation search space
   \STATE All\_samples = Randomly select samples from Pool
   \STATE Train (All\_samples)
   \STATE Predictor = Build a GP predictor with all observations
   \WHILE {Eval(Predictor) $\geq$ $e$}
   \FOR {Architecture in (Pool $\setminus$ All\_samples)}
   \STATE $\rm{a}_{i}$ = {Predictor}.$\textit{getAccuracy}$(Architecture)
   \STATE $\rm{v}_{i}$ = {Predictor}.$\textit{getUncertainty}$(Architecture)
   \STATE $\rm{flop}_{i}$ = $\textit{getFLOP}$(Architecture)
   \ENDFOR
   \STATE S$_{\rm{explore}}$ = \{Architectures with $p$ highest $\rm{v}_{i}$\}
   \STATE S$_{\rm{exploit}}$ = \{Architectures with $q$ highest $\frac{\rm{a}_{i}}{\rm{flop}_{i}}$\}
   \STATE All\_samples = All\_samples $\cup$ S$_{\rm{explore}}$ $\cup$ S$_{\rm{exploit}}$
   \STATE Train (S$_{\rm{explore}}$ $\cup$ S$_{\rm{exploit}}$)
   \STATE Predictor = Build a GP predictor with all observations
   \ENDWHILE
   \STATE \textbf{Return} Predictor
\end{algorithmic}
\end{algorithm}

\subsubsection{Iterative Sample Selection}

As mentioned earlier, our objective is to train the GP predictor with as few 
NN architecture samples as possible.  We summarize our efficient sample 
generation and predictor training method in Algorithm~\ref{alg:accu_pred}. 
Since the number of unique architectures in the adaptation search space can 
still be large, we first sample representative architectures from this search 
space to form an architecture pool.  We adopt the QMC sampling 
method~\cite{qmc}, which is known to provide similar accuracy to Monte Carlo 
sampling but with orders of magnitude fewer samples.  We then build the 
accuracy predictor iteratively.  In each iteration, we use the current 
predictor as a guide for selecting additional sample architectures to add to 
the training set.  We train these sample architectures and then upgrade the 
predictor based on new architecture-accuracy observations. 

\begin{figure}[t]
\begin{center}
\includegraphics[width=69mm]{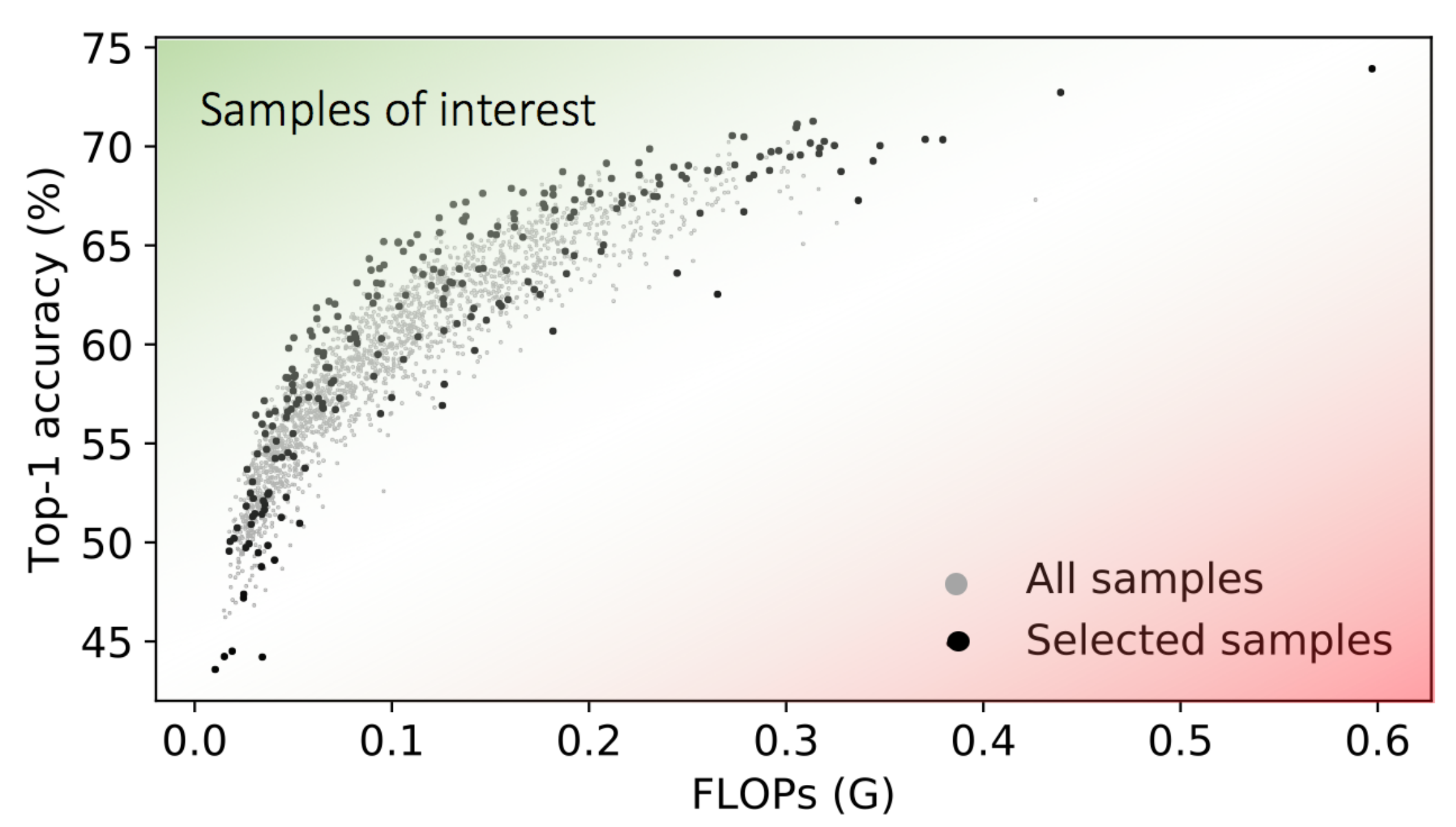}
\caption{An illustration of `samples of interest' and sample selection result.}
\label{fig:bayesian_op}
\end{center}
\vspace{-1mm}
\end{figure}

To improve sample efficiency, we further incorporate Bayesian optimization 
into the sample architecture selection process. This enables 
Algorithm~\ref{alg:accu_pred} to converge faster with fewer 
samples~\cite{bayesian_op}.  Specifically, we select both exploitation and 
exploration samples in each iteration:
\begin{itemize}
    \item Exploitation samples: We choose sample architectures with high 
accuracy/FLOPs ratios.  These desirable architectures, or `samples of
interest,' are likely to yield higher accuracy with less computation
cost.  They typically fall in the top left part of the accuracy-FLOPs 
trade-off graph, as shown in Fig.~\ref{fig:bayesian_op}. 
    \item Exploration samples: We choose samples with large uncertainty 
values.  This helps increase the prediction confidence level of the GP 
regressor over the entire adaptation search space~\cite{bayesian_op}.
\end{itemize}
Based on these rules, we show the selected sample architectures from the 
architecture space in Fig.~\ref{fig:bayesian_op}.  It can be observed that 
we have higher sampling density in the area of `samples of interest,' where 
adaptation typically rests.

\subsection{Latency Predictor}
\begin{figure}[t]
\begin{center}
\begin{tabular}{cc}
\includegraphics[width=41mm]{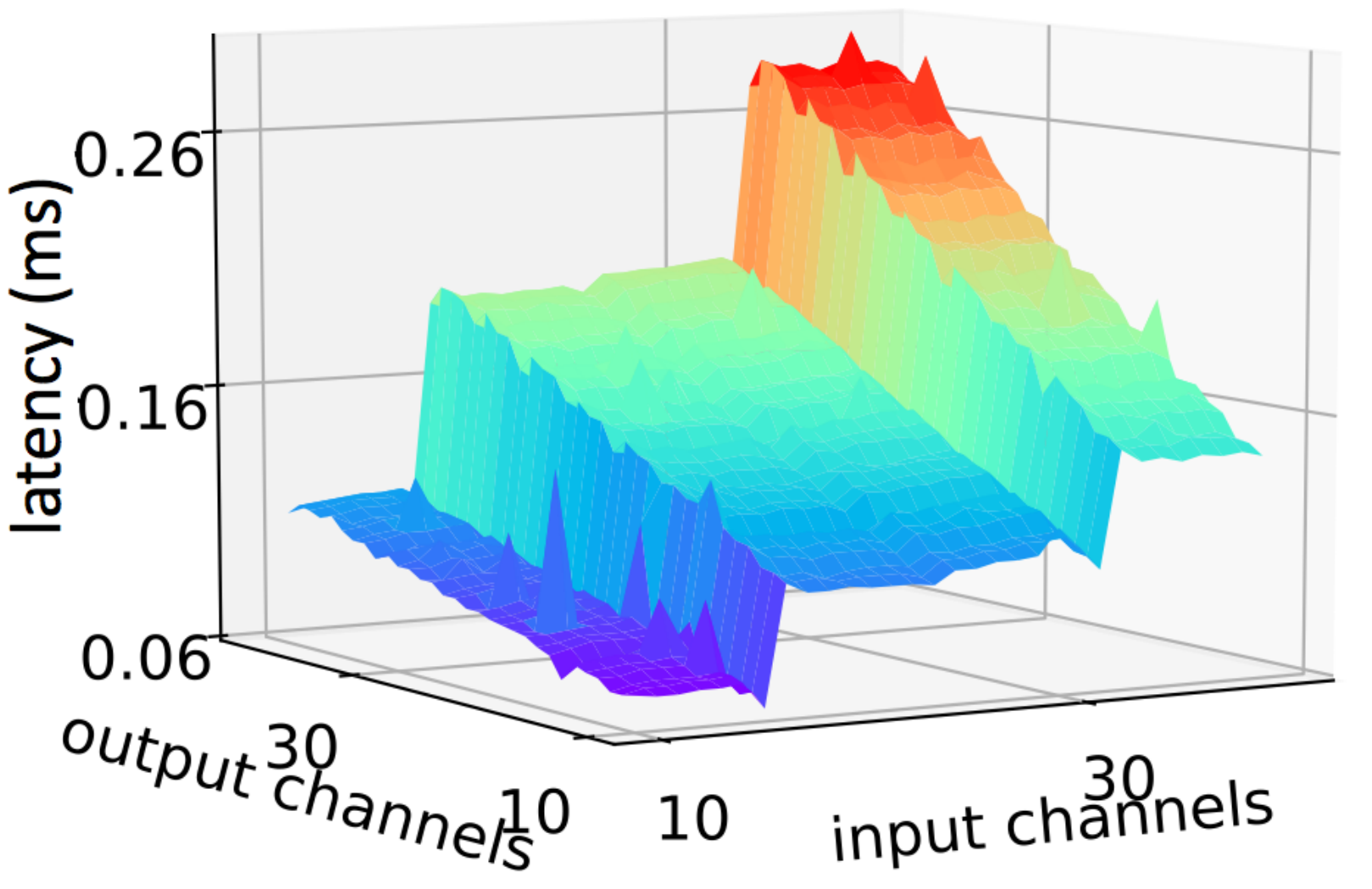}
&
\includegraphics[width=41mm]{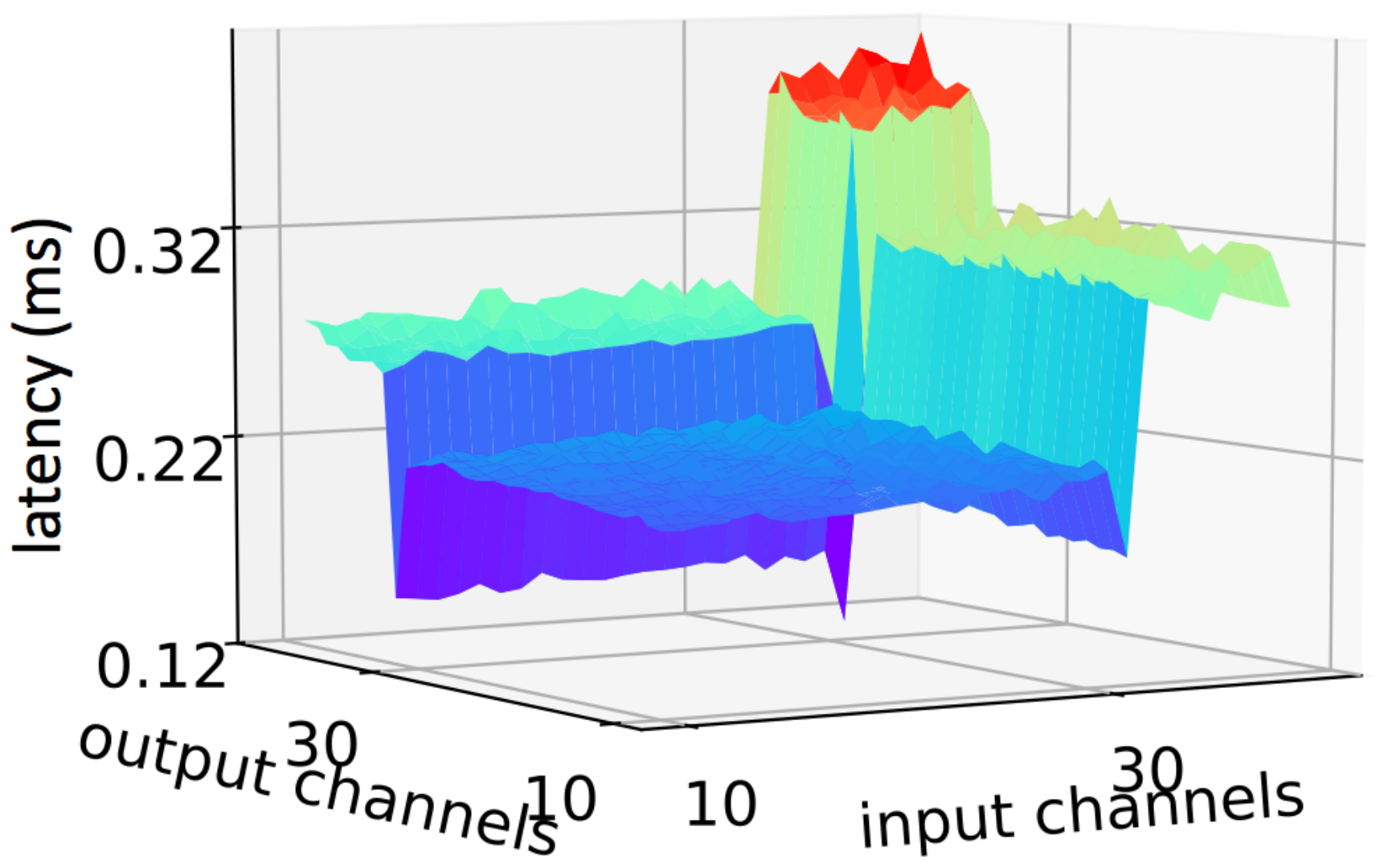}
\\
(a) & (b)
\\
\end{tabular}
\caption{Latency vs. \#Channels for a 1$\times$1 convolution on an input 
image size of 56$\times$56 and stride 1 on (a) Snapdragon 835 GPU and
(b) Hexagon v62 DSP.}
\label{fig:latency_flop}
\end{center}
\vspace{-2mm}
\end{figure}

Recently, great efforts have been made towards developing more efficient and 
compact NN architectures for better accuracy-latency trade-offs.  Most of them 
optimize NNs based on FLOPs, which is an often-used 
proxy~\cite{nasnet, nas1}.  However, optimization based on direct latency 
measurement instead of FLOPs can better explore hardware traits and hence 
offer additional advantages.  To illustrate this point, we show the measured 
latency surface of a 1$\times$1 convolution operator with varying numbers of 
input and output channels in Fig.~\ref{fig:latency_flop}.  The latency is 
measured using the Caffe2 library on a Snapdragon 835 mobile CPU and a 
Hexagon v62 DSP.  It can be observed that FLOPs, though generally effective 
in providing guidance for latency reduction, may not capture desired hardware 
characteristics upon model deployment.

Extracting the latency of an NN architecture during EES execution through direct 
measurement, however, is very challenging.  Platform-specific latency 
measurements can be slow and difficult to parallelize, especially when the
 number of available devices is limited~\cite{netadapt}.  Therefore, 
large-scale latency measurements might be expensive and become the computation 
bottleneck for Chameleon.  To speed up this process, we construct an operator 
latency LUT for the target device to enable fast and reliable latency 
estimations of NN candidates throughout EES execution.  The LUT is supported 
by an operator latency database, where we benchmark operator-level latency on 
real devices with different input dimensions.  For an NN model, we sum up all its 
operator-level latencies as an estimate of the network-level latency:
\begin{equation}
    t_{net} = \Sigma\ t_{operator}
\end{equation}
Building the operator latency LUT for a given device is just a one-time cost, 
but can be substantially reused across various NN models, different tasks, 
and different applications of architecture search, model adaptation, and 
hyperparameter optimization.

\begin{figure}[t]
\begin{center}
\includegraphics[width=55mm]{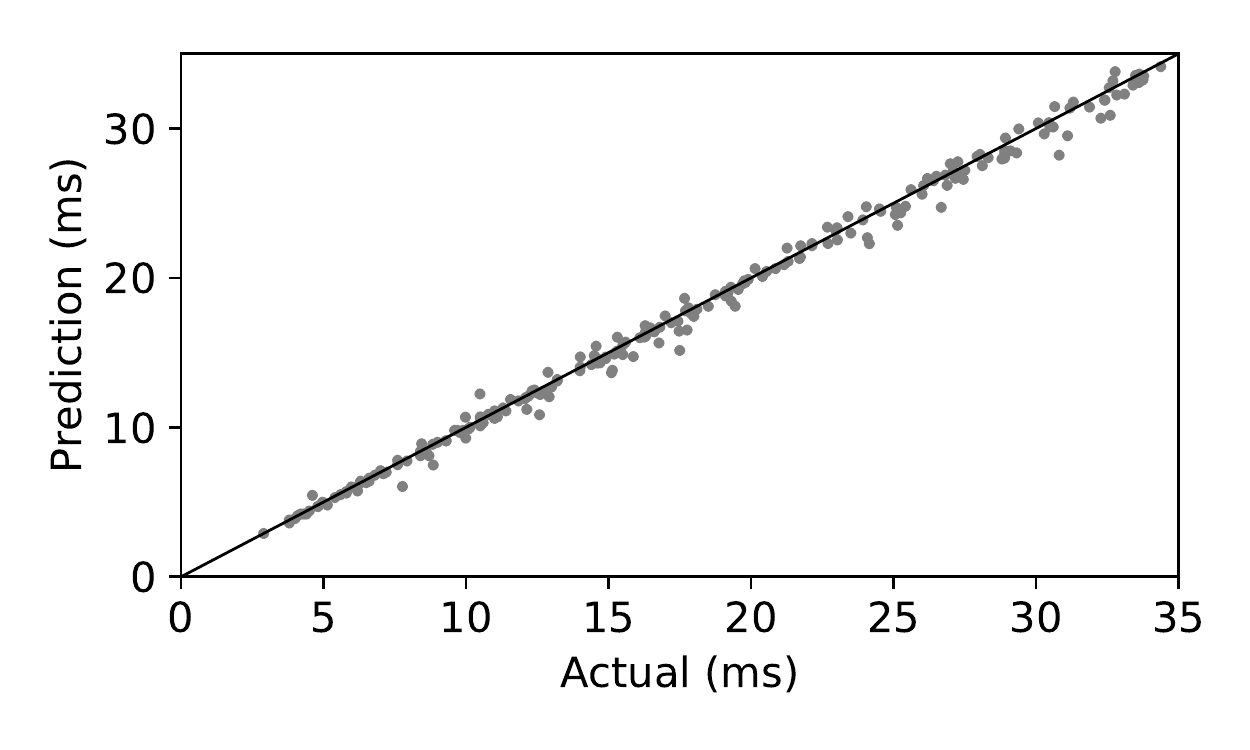}
\caption{LUT-based latency predictor evaluation on Snapdragon 835 CPU.}
\label{fig:latency_energy_pred}
\end{center}
\vspace{-2mm}
\end{figure}

Latency estimation based on operator latency LUT can be completed in less 
than one CPU second, as opposed to real measurements on hardware that usually 
take minutes.  Moreover, it also supports parallel query, hence significantly 
enhancing simultaneous latency extraction efficiency across multiple NN 
candidates.  This enables latency estimation in EES to consume very
little time.  
We compare the predicted latency value against real 
measurement in Fig.~\ref{fig:latency_energy_pred}.  The distance between 
the predicted value and real measurement is quite small.  We plan to 
open-source the operator latency LUT for mobile CPUs and DSPs.

\subsection{Energy Predictor}
Battery-powered devices (e.g., smart watches, mobile phones, and AR/VR 
products) have limited energy budgets.  Thus, it is important to adjust and 
fine-tune the NN model to fit the energy constraint before 
deployment~\cite{neuralpower}.  To solve this problem, we incorporate 
energy constraint-driven adaptation in Chameleon for different platforms and 
use scenarios.

\begin{figure}[h]
\begin{center}
\includegraphics[width=55mm]{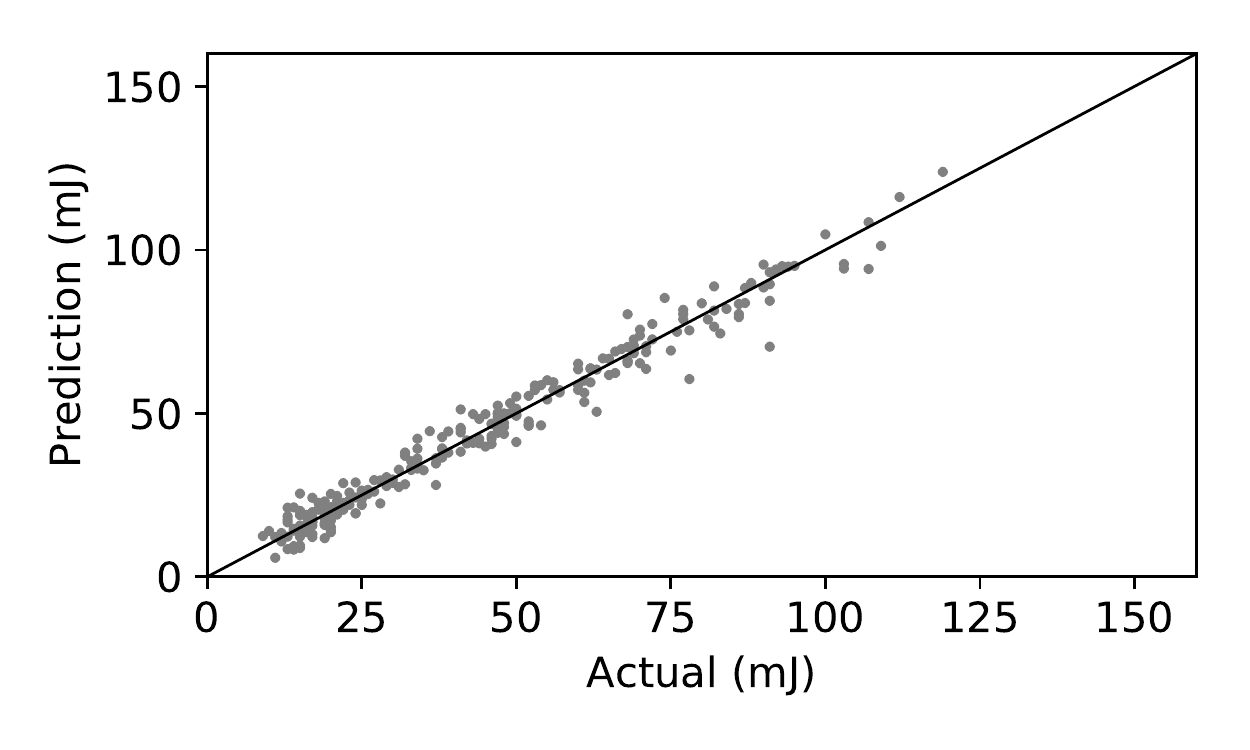}
\caption{Energy predictor evaluation on Snapdragon 835 CPU.}
\label{fig:energy_pred}
\end{center}
\vspace{-2mm}
\end{figure}

We build the energy predictor in a similar manner to the accuracy predictor.  
We build a GP energy predictor that incorporates Bayesian optimization and
acquire the energy values from direct measurements on hardware.  However, 
for the energy predictor, we only select exploration samples in each iteration 
(i.e., samples with large uncertainty).  The concept of `samples of interest' 
is not applicable in this scenario.  
We show the performance of our energy predictor for MobileNetV2 in 
Fig.~\ref{fig:energy_pred}.  

\section{Experiments}
In this section, we apply Chameleon to various NN architectures over a wide 
spectrum of platforms and resource budgets.  We use PyTorch~\cite{pytorch} and 
Caffe2 for the implementation.  For mobile platforms, we train the model with full precision (float32) and quantize to int8.  We report full precision results for accuracy comparison but we see no or minimal loss of accuracy for quantized models with fake- quantization fine-tuning.  We benchmark the latency using Caffe2 int8 back-end with Facebook AI Performance Evaluation Platform~\cite{aibench}.
We report results 
on the ImageNet dataset~\cite{imagenet}, which is a well-known benchmark 
consisting of 1.2M training and 50K validation images classified into 1000 distinct classes.  
Thus, all our models share the same 1000 final output channels.  We randomly reserve 50K images from the training set (50 images per class) to build the accuracy predictor.  We measure 
latency and energy with the same batch size of 1.  

In the EES process, we set $\alpha=10/\rm{ms}$ and $\alpha=10/\rm{mJ}$ for 
latency- and energy-constrained adaptation, respectively, and $w=2$.  We set 
the initial QMC architecture pool size to $k=2048$.  We generate the accuracy 
and energy predictors with 240 samples selected from the architecture pool.  
Latency estimation is supported by a operator latency LUT with approximately 
350K records.  In evolutionary search, the population size of each generation 
is set to 96. We pick the top 12 candidates for the next generation.  The 
total number of search iterations is set to 100.  We present our experimental 
results next.

\subsection{Adaptation for Mobile Models}
This section presents the adaptation results leveraging the efficient inverse residual building block from MobileNetV2, which is the state-of-the-art hand-crafted architecture for mobile platforms.  It utilizes 
inverted residual and linear bottleneck to significantly cut down on the 
number of operations and memory needed per inference~\cite{mobilenetv2}.  We 
first show the adaptation search space used in our experiments in Table~\ref{tab:MobileNetV2_range}, where $t$, $c$, $n$, and 
$s$ refer to the expansion factor, number of output channels, number of 
repeated blocks, and stride, respectively.  The adaptation search range for 
each hyperparameter is denoted as [$a, b$], where $a$ denotes the lower bound 
and $b$ the upper bound.  The default values used in MobileNetV2 1.0x are 
also shown next to our search ranges, following the notation rule 
used in~\cite{mobilenetv2}.

\begin{table}[h]
\centering
\small
\caption{Adaptation space of ChamNet-Mobile}
\label{tab:MobileNetV2_range}
\begin{tabular}{l|c|c|c|c}
\hline
\multicolumn{2}{l}{Input resolution $\rightarrow$} & \multicolumn{2}{l}{224 [96, 224]} \\
\hline
stage & $t$ & $c$ & $n$ & $s$\\
\hline
conv2d & - & 32 [8,48] & 1 & 2\\
\hline
bottleneck & 1 & 16 [8,32] & 1 & 1 \\
\hline
bottleneck & 6 [2,6] & 24 [8,40] & 2 [1,2] & 2 \\
\hline
bottleneck & 6 [2,6] & 32 [8,48] & 3 [1,3]& 2 \\
\hline
bottleneck & 6 [2,6] & 64 [16,96] & 4 [1,4]& 2 \\
\hline
bottleneck & 6 [2,6] & 96 [32,160]& 3 [1,3]& 1 \\
\hline
bottleneck & 6 [2,6] & 160 [56,256] & 3 [1,3]& 2 \\
\hline
bottleneck & 6 [2,6] & 320 [96,480] & 1 & 1 \\
\hline
conv2d & - & 1280 [1024,2048] & 1 & 1 \\
\hline
avgpool & - & - & 1 & - \\
\hline
fc & - & 1000 & - & - \\
\hline
\end{tabular}
\end{table}

\begin{figure}[t]
\begin{center}
\includegraphics[width=72mm]{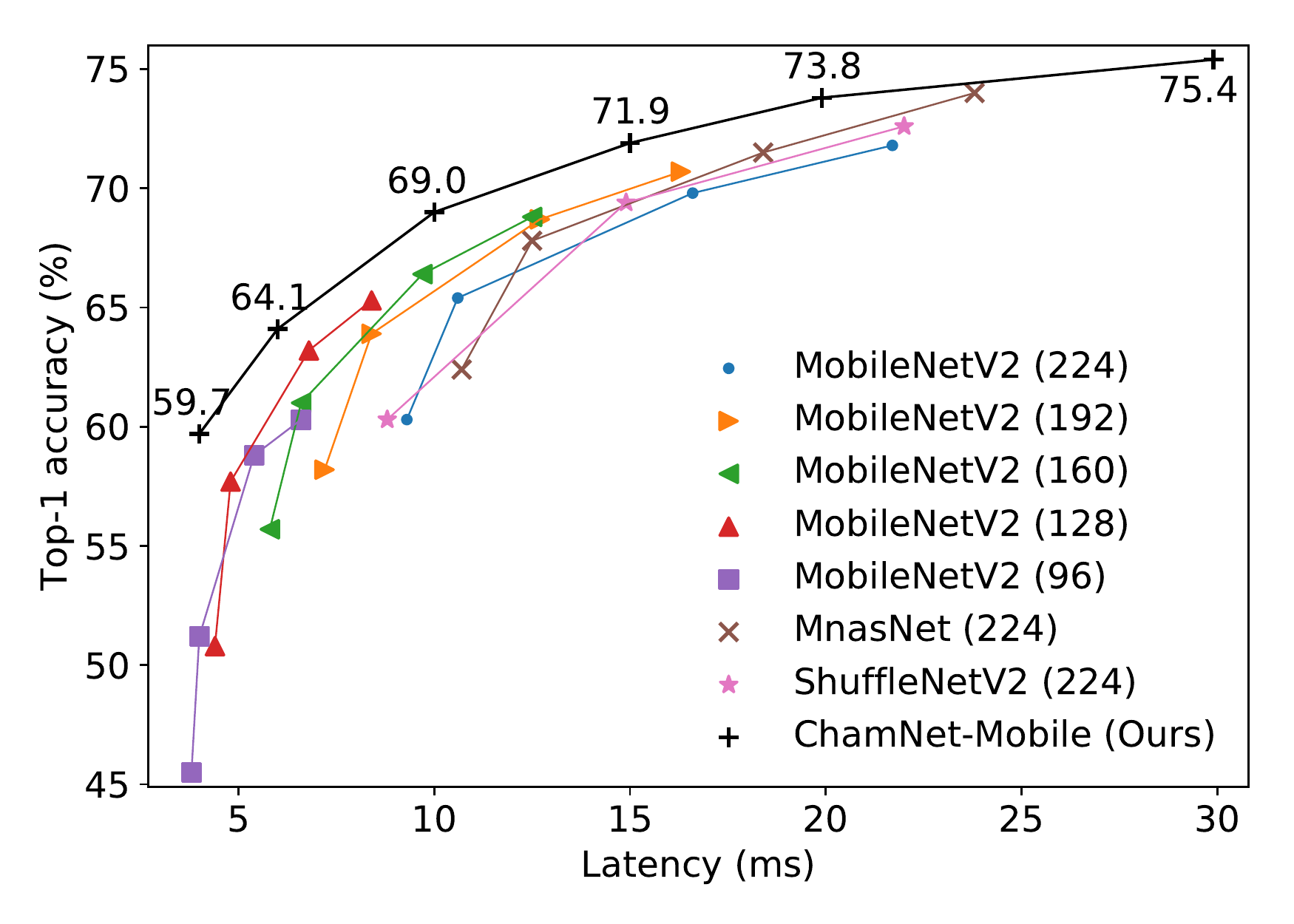}
\caption{Performance of ChamNet-Mobile on a Snapdragon 835 CPU.  Numbers in 
parentheses indicate input image resolution.}
\label{fig:mobilenet_v2_mobile_CPU}
\end{center}
\vspace{-2mm}
\end{figure}

\begin{figure}[t]
\begin{center}
\includegraphics[width=72mm]{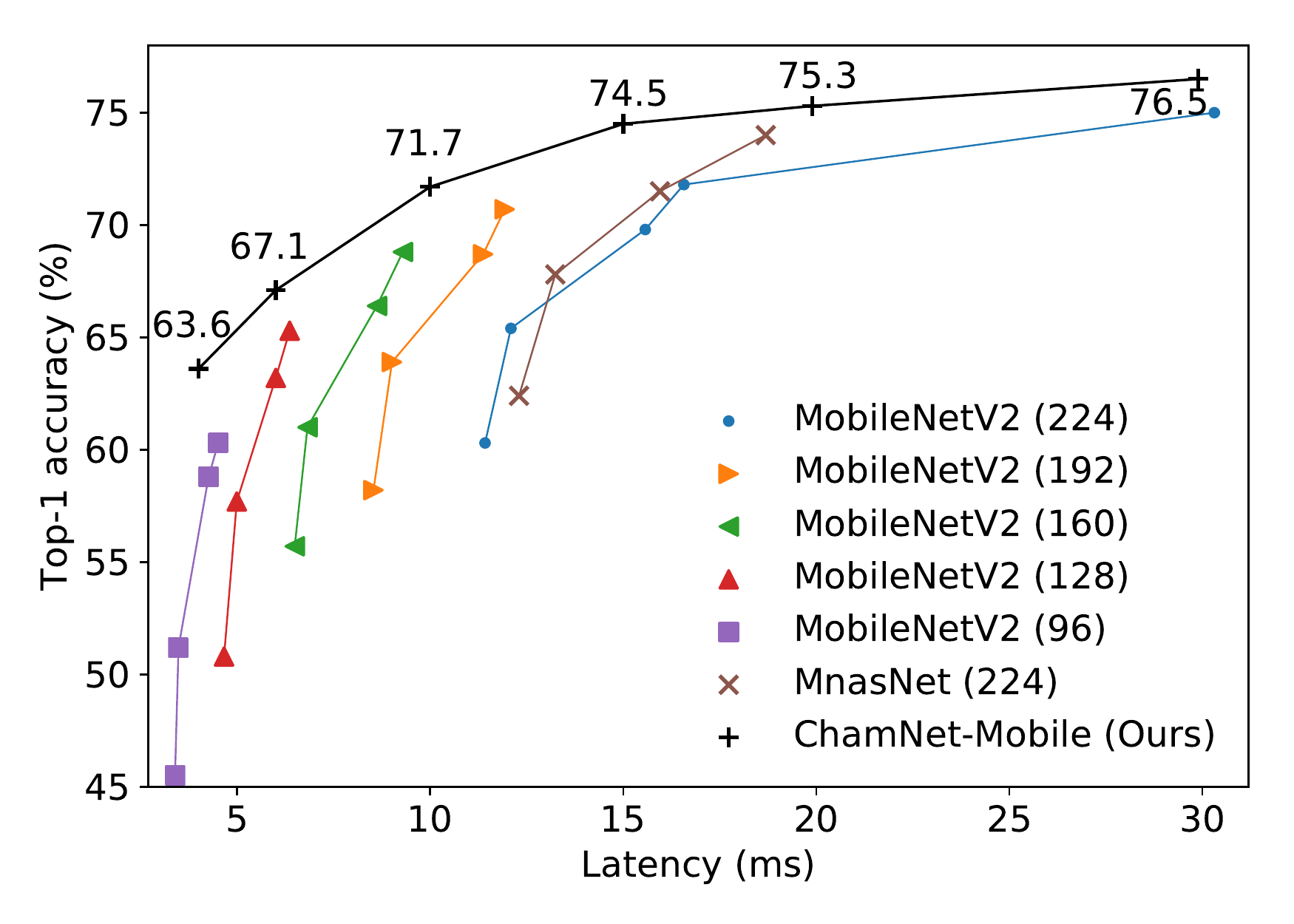}
\caption{Performance of ChamNet-Mobile on a Hexagon v62 DSP. Numbers in 
parentheses indicate input image resolution.}
\label{fig:mobilenet_v2_mobile_dsp}
\end{center}
\vspace{-2mm}
\end{figure}

We target two different platforms in our experiments: Snapdragon 835 mobile CPU (on a Samsung S8) and Hexagon v62 DSP 
(800 MHz frequency with internal NN library implementation).  We evaluate Chameleon under a wide 
range of latency constraints: 4ms, 6ms, 10ms, 15ms, 20ms, and 30ms.

We compare our adapted ChamNet-Mobile models with state-of-the-art models, including MobileNetV2, 
ShuffleNetV2, and MnasNet in Fig.~\ref{fig:mobilenet_v2_mobile_CPU} and 
\ref{fig:mobilenet_v2_mobile_dsp} on mobile CPU and DSP, respectively.  Models discovered by 
Chameleon outperform all the previous manually designed or automatically searched architectures 
on both platforms consistently.  Our ChamNet has an 8.5\% absolute accuracy gain compared 
to MobileNetV2 0.5x with an input resolution of 96$\times$96, while both models share 
the same 4.0ms run-time latency on the mobile CPU.

\subsection{Adaptation for Server Models}
We also evaluate Chameleon for server models on both CPU and GPU.  We choose residual building
blocks from ResNet as the base for adaptation because of its high accuracy and widespread usage in the cloud. The target platforms are 
the Intel Xeon Broadwell CPU with 2.4 GHz frequency and Nvidia GTX 1060 GPU with 1.708 GHz
frequency. We use CUDA 8.0 and CUDNN 5.1 in our experiments.

We show the detailed adaptation search space in Table~\ref{tab:resnet_range}, where the notations are 
identical to the ones in Table~\ref{tab:MobileNetV2_range}.  This search space for ChamNet-Res includes 
\#Filters, expansion factor, and \#Bottlenecks per layer.  Note that the maximum number of layers in 
the adaptation search space is 152, which is the largest reported depth for the ResNet 
family~\cite{resnet}.

\begin{table}[t]
\centering
\caption{Adaptation space of ChamNet-Res}
\label{tab:resnet_range}
\small
\begin{tabular}{l|c|c|c|c}
\hline
\multicolumn{2}{l}{Input resolution $\rightarrow$} & \multicolumn{2}{l}{224} \\
\hline
stage & t & c & n & s\\
\hline
conv2d & - & 64 [16,64] & 1 & 2 \\
\hline
bottleneck & 4 [2,6] & 64 [16,64] & 3 [1,3] & 2 \\
\hline
bottleneck & 4 [2,6] & 128 [32,128] & 4 [1,8] & 2 \\
\hline
bottleneck & 4 [2,6] & 256 [64,256] & 6 [1,36] & 2 \\
\hline
bottleneck & 4 [2,6] & 512 [128,512] & 3 [1,3] & 2 \\
\hline
avgpool & - & - & 1 & - \\
\hline
fc & - & 1000 & - & -\\
\hline
\end{tabular}
\end{table}

\begin{figure}[t]
\begin{center}
\includegraphics[width=72mm]{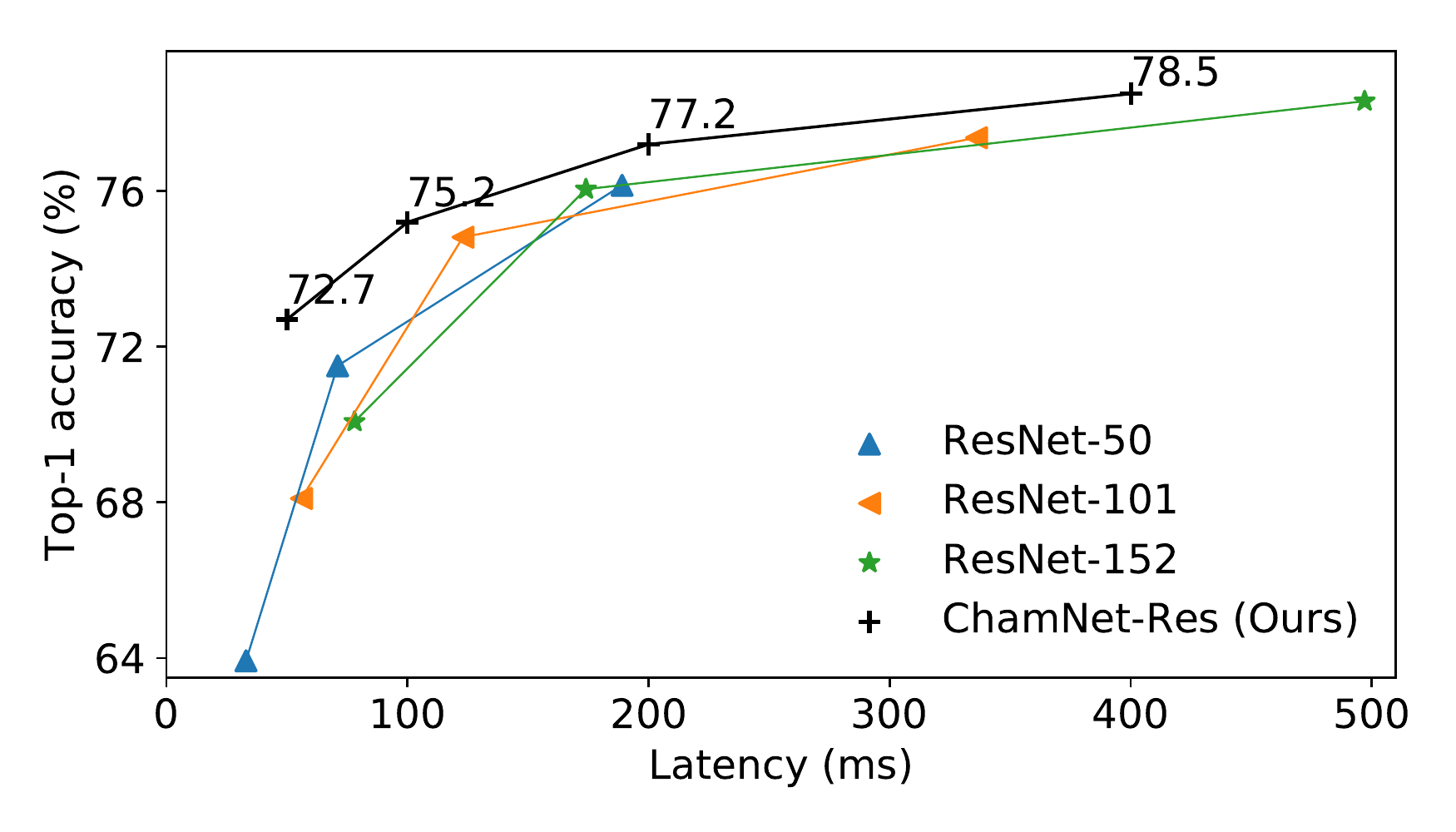}
\caption{Latency-constrained ChamNet-Res on an Intel CPU.}
\label{fig:resnet_cpu}
\end{center}
\vspace{-3mm}
\end{figure}

\begin{figure}[t]
\begin{center}
\includegraphics[width=72mm]{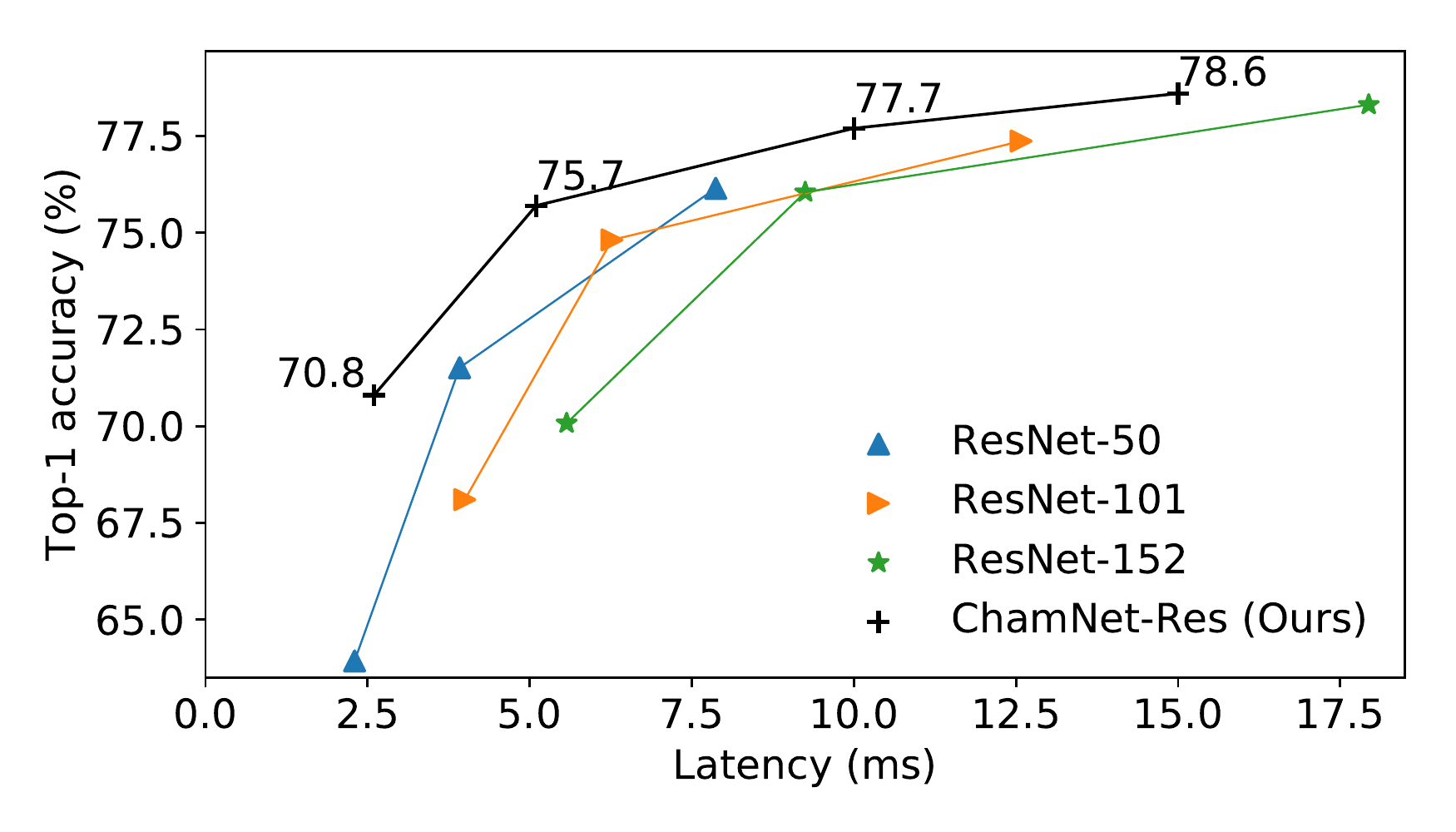}
\caption{Latency-constrained ChamNet-Res on an Nvidia GPU.}
\label{fig:resnet_gpu}
\end{center}
\vspace{-3mm}
\end{figure}

We set a wide spectrum of latency constraints for both Intel CPU and Nvidia GPU to demonstrate the 
generality of our framework.  The latency constraints for the CPU are 50ms, 100ms, 200ms, and 400ms, 
while the constraints for the GPU are 2.5ms, 5ms, 10ms, and 15ms.  We compare the adapted model with 
ResNets on CPU and GPU in Fig.~\ref{fig:resnet_cpu} 
and~\ref{fig:resnet_gpu}, respectively.  Again, Chameleon improves the accuracy by a large margin on 
both platforms.

\subsection{Energy-Driven Adaptation}
Next, we study the energy-constrained use scenario for ChamNet-Mobile on mobile phones.  We
obtain the energy measurements from the Snapdragon 835 CPU.  We first replace the battery of the 
phone with a Monsoon power monitor with a constant voltage output of 4.2V.  During measurements, we 
ensure the phone is kept in the idle mode for 18 seconds, then run the network 1000 times and measure 
the current at 200$\mu$s intervals.  
We then deduct the baseline current from the raw data in a post-processing step and 
calculate the energy consumption per forward pass.

To demonstrate Chameleon's applicability under a wide range of constraints, we set 
six different energy constraints in our experiment: 15mJ, 30mJ, 50mJ, 75mJ, 100mJ, and 150mJ. We use the 
same adaptation search space as ChamNet-Mobile.

\begin{figure}[t]
\begin{center}
\includegraphics[width=75mm]{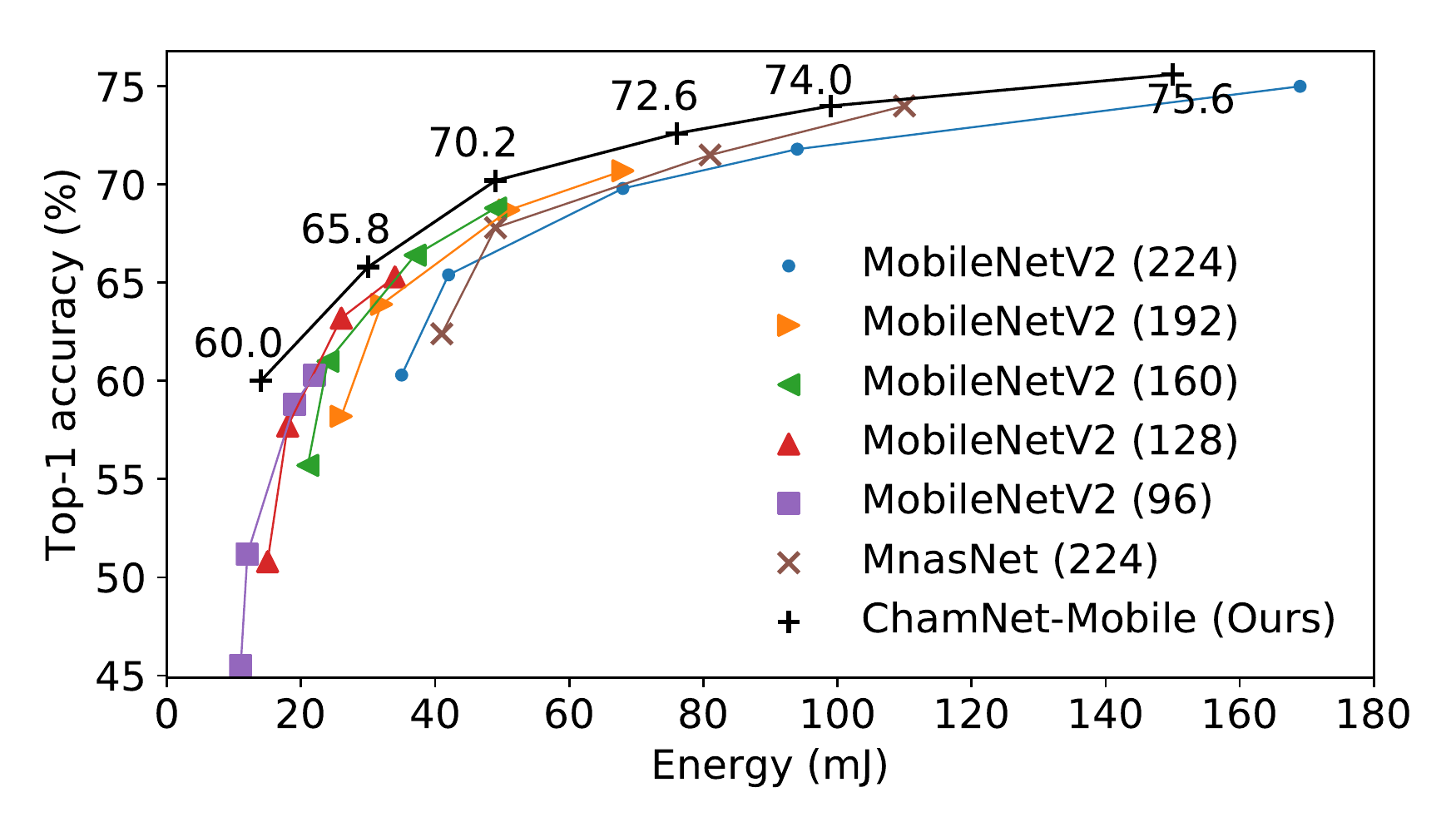}
\caption{Energy-constrained ChamNet-Mobile on a Snapdragon 835 CPU. Numbers in 
parentheses indicate input image resolution.}
\label{fig:mobilenet_v2_mobile_CPU_energy}
\end{center}
\vspace{-2mm}
\end{figure}

We show the accuracy of ChamNet and compare it with MobileNetV2 in 
Fig.~\ref{fig:mobilenet_v2_mobile_CPU_energy}.  We achieve significant improvement in 
accuracy-energy trade-offs.  For example, compared to the MobileNetV2 0.75x with input resolution 96$\times$96 baseline 
(58.8\% accuracy at 19mJ per run), our adapted models achieves 60.0\% accuracy at only 14mJ 
per run.  Therefore, our model is able to reduce energy by 26\% while simultaneously increasing
accuracy by 1.2\% on a smartphone.

\begin{table*}[h!]
\centering
\caption{Comparisons of different architecture search and model adaptation approaches.}
\resizebox{150mm}{!}{%
\small
\begin{tabular}{lccccc}
\hline
Model & Method  & Direct & Scaling & \ \ \ \ \ Latency(ms) \ \ & Top-1 \\
& & metrics based & \ \ complexity\ \  & & accuracy (\%)\\
\hline
MobileNetV2 1.3x~\cite{mobilenetv2}\ \ \ &  Manual\  & $-$\  & $-$\  & 33.8\  & 74.4\\
ShuffleNetV2 2.0x~\cite{shufflenetv2} & Manual & Y & $-$ & 33.3 & 74.9\\
\textbf{ChamNet-A} & \textbf{EES}  & \textbf{Y} &\textbf{$\mathcal{O} (m+n)$} &\textbf{29.8} & \textbf{75.4} \\
\hline
MobileNetV2 1.0x~\cite{mobilenetv2}\ \ \ &  Manual\  & $-$\  & $-$\  & 21.7\  & 71.8\\
ShuffleNetV2 1.5x~\cite{shufflenetv2} & Manual & Y & $-$ & 22.0 & 72.6\\
CondenseNet (G=C=4)~\cite{condensenet} & Manual & $-$ & $-$ & 28.7$^{*}$ & 73.8 \\
MnasNet 1.0x~\cite{mnasnet} & RL  & Y & $\mathcal{O} (m\cdot n\cdot k)$ & 23.8 & \textbf{74.0} \\
AMC~\cite{amc} & RL  & Y & $\mathcal{O} (m\cdot n\cdot k)$ & $-$ & $-$\\
MorphNet~\cite{morphnet} & Regularization & N & $\mathcal{O} (m\cdot n\cdot k)$ & $-$ & $-$\\
\textbf{ChamNet-B} & \textbf{EES}  & \textbf{Y} &\textbf{$\mathcal{O} (m+n)$} &\textbf{19.9} & 73.8 \\
\hline
MobileNetV2 0.75x~\cite{mobilenetv2}\ \ \ &  Manual\  & $-$\  & $-$\  & 16.6\  & 69.8\\
ShuffleNetV2 1.0x~\cite{shufflenetv2} & Manual  & Y & $-$ & \textbf{14.9} & 69.4\\
MnasNet 0.75x~\cite{mnasnet} & RL  & Y & $\mathcal{O} (m\cdot n\cdot k)$ & 18.4 & 71.5 \\
NetAdapt~\cite{netadapt} & Pruning  & Y & $\mathcal{O} (m\cdot k + n)$ &  16.6 (63.6$^{+}$) & 70.9 \\
\textbf{ChamNet-C} & \textbf{EES}  &  \textbf{Y} & \textbf{$\mathcal{O} (m+n)$} &15.0 & \textbf{71.9} \\
\hline
MobileNetV2 0.5x~\cite{mobilenetv2}& Manual & $-$ & $-$ & 10.6 & 65.4\\
MnasNet 0.35x~\cite{mnasnet} & RL  & Y & $\mathcal{O} (m\cdot n\cdot k)$ & 10.7 & 62.4 \\
\textbf{ChamNet-D} & \textbf{EES}  &  \textbf{Y} & \textbf{$\mathcal{O} (m+n)$} &\textbf{10.0} & \textbf{69.0} \\
\hline
MobileNetV2 0.35x~\cite{mobilenetv2}\ \ \ &  Manual\  & $-$\  & $-$\  & 9.3\  & 60.3\\
ShuffleNetV2 0.5x~\cite{shufflenetv2} & Manual  & Y & $-$ & 8.8 & 60.3\\
\textbf{ChamNet-E} & \textbf{EES}  & \textbf{Y} &\textbf{$\mathcal{O} (m+n)$} &\textbf{6.1} & \textbf{64.1} \\

\hline
\multicolumn{6}{l}{\scriptsize{We report five of our ChamNet models with A-30ms, B-20ms, C-15ms, D-10ms, and E-6ms latency constraints. $m$, $n$, and $k$ refer to the number of network }}\\
\multicolumn{6}{l}{\scriptsize{models, distinct platforms, and use scenarios with different resource budgets, respectively. $^{+}$: Ref.~\cite{netadapt} reports 63.6ms latency with TensorFlow Lite on Pixel 1. }}\\
\multicolumn{6}{l}{\scriptsize{For a fair comparison, we report the corresponding latency in  our experimental setup with Caffe2 on Samsung Galaxy S8 with Snapdragon 835 CPU. $^{*}$: The}}\\
\multicolumn{6}{l}{\scriptsize{inference engine is faster than other models.}}\\
\end{tabular}%
}
\vspace{-2mm}
\label{tab:algo_comparison}
\end{table*}

\subsection{Comparisons with Alternative Adaptation and Compression Approaches}

In this section, we compare Chameleon with relevant work, including:

\begin{enumerate}\itemsep-0em 
\item MNAS~\cite{mnasnet}: this is an RL-based NN architecture search algorithm for mobile devices.
\item AutoML model compression (AMC)~\cite{amc}: this is an RL-based automated network compression method.
\item NetAdapt~\cite{netadapt}: this is a platform-aware filter pruning algorithm that
adapts a pre-trained network to a specific hardware under a given latency constraint.
\item MorphNet~\cite{morphnet}: this is a network simplification algorithm based on sparsifying 
regularization.
\end{enumerate}

Table~\ref{tab:algo_comparison} compares different model compression, adaptation, and optimization 
approaches on the Snapdragon 835 CPU, where $m$, $n$, and $k$ refer to the number of network models, distinct platforms, and use scenarios with different resource budgets, respectively.  ChamNet yields the most 
favorable accuracy-latency trade-offs among all models.  Moreover, most existing approaches need to 
be executed at least once per network per device per constraint~\cite{mnasnet, netadapt, morphnet, amc}, and 
thus have a total training cost of $\mathcal{O} (m\cdot n\cdot k)$.  Chameleon only builds
$m$ accuracy predictors and $n$ resource predictors (e.g., latency LUT), and thus reduces the cost 
to $\mathcal{O} (m+n)$. The search cost is negligible once the predictors are built.  Such one-time costs can easily be amortized when the number of use scenarios scales up, which is generally 
the case for large-scale heterogeneous deployment.

\begin{figure}[t]
\begin{center}
\includegraphics[width=70mm]{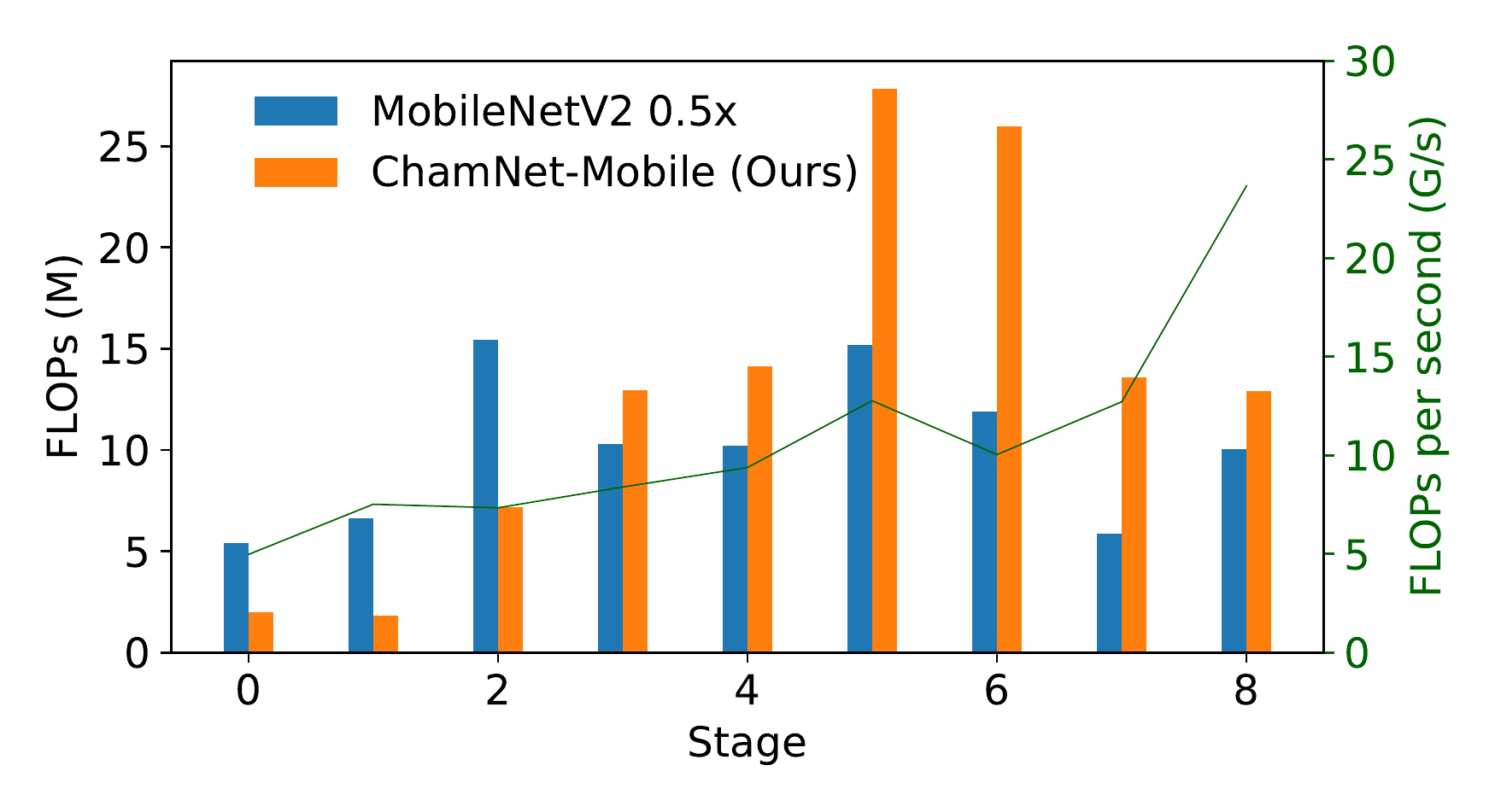}
\caption{FLOPs distribution and stage-wise CPU processing speed of MobileNetV2 and ChamNet on a mobile CPU.}
\label{fig:flop_compare}
\end{center}
\vspace{-2mm}
\end{figure}

We compare the FLOPs distribution at each stage (except for avgpool and fc) for 
MobileNetV2 0.5x and ChamNet with similar latency in Fig.~\ref{fig:flop_compare}.  
Our model achieves 71.9\% accuracy at 15.0ms compared to the MobileNetV2
that has 69.8\% accuracy at 16.6ms.  We have two observations:
\begin{enumerate}\itemsep-0em
\item ChamNet redistributes the FLOPs from the early stages to late stages.  We 
hypothesize that this is because when feature map size is smaller in the later stages, more filters 
or a larger expansion factor are needed to propagate the information.

\item ChamNet has a better utilization of computation resources. We estimate the CPU 
processing speed at each stage using the ratio of FLOPs and latency, as shown with the green curve 
in Fig.~\ref{fig:flop_compare}.  The operators in early stages with large input image size have 
significantly lower FLOPs per second, hence incur higher latency given the same computational load.  
A possible reason is incompatibility between cache capacity and large image size.  Through better FLOPs
redistribution, ChamNet enables 2.1\% accuracy gain while reducing run-time 
latency by 5\% against the baseline MobileNetV2.
\end{enumerate}

\section{Conclusions}
This paper proposed a platform-aware model adaptation framework called Chameleon that leverages
efficient building blocks to adapt a model to different real-world platforms and use scenarios.  
This framework is based on 
very efficient predictive models and thus bypasses the expensive training and measurement process.  
It significantly improves accuracy without incurring any latency or energy overhead, while 
taking only CPU minutes to perform an adaptation search.  At the same latency or energy, 
it achieves significant accuracy gains relative to both handcrafted and automatically searched models.


{\small
\bibliographystyle{ieee}
\bibliography{bib}

\begin{thebibliography}{10}\itemsep=-1pt

\bibitem{aibench}
Facebook {AI} performance evaluation platform.
\newblock \url{https://github.com/facebook/FAI-PEP}, 2018.

\bibitem{qmc}
S.~Asmussen and P.~W. Glynn.
\newblock {\em {Stochastic Simulation: Algorithms and Analysis}}, volume~57.
\newblock Springer Science \& Business Media, 2007.

\bibitem{archsearch}
B.~Baker, O.~Gupta, R.~Raskar, and N.~Naik.
\newblock Accelerating neural architecture search using performance prediction.
\newblock {\em arXiv preprint arXiv:1705.10823}, 2017.

\bibitem{smboalgo}
J.~S. Bergstra, R.~Bardenet, Y.~Bengio, and B.~K{\'e}gl.
\newblock Algorithms for hyper-parameter optimization.
\newblock In {\em Proc. Advances in Neural Information Processing Systems},
  pages 2546--2554, 2011.

\bibitem{neuralpower}
E.~Cai, D.-C. Juan, D.~Stamoulis, and D.~Marculescu.
\newblock Neuralpower: {Predict} and deploy energy-efficient convolutional
  neural networks.
\newblock {\em arXiv preprint arXiv:1710.05420}, 2017.

\bibitem{NeST}
X.~Dai, H.~Yin, and N.~K. Jha.
\newblock {NeST}: {A} neural network synthesis tool based on a grow-and-prune
  paradigm.
\newblock {\em arXiv preprint arXiv:1711.02017}, 2017.

\bibitem{lstmprune}
X.~Dai, H.~Yin, and N.~K. Jha.
\newblock Grow and prune compact, fast, and accurate {LSTMs}.
\newblock {\em arXiv preprint arXiv:1805.11797}, 2018.

\bibitem{imagenet}
J.~Deng, W.~Dong, R.~Socher, L.-J. Li, K.~Li, and L.~Fei-Fei.
\newblock {ImageNet}: {A} large-scale hierarchical image database.
\newblock In {\em Proc. IEEE Conf. Computer Vision and Pattern Recognition},
  pages 248--255. Ieee, 2009.

\bibitem{morphnet}
A.~Gordon, E.~Eban, O.~Nachum, B.~Chen, H.~Wu, T.-J. Yang, and E.~Choi.
\newblock {MorphNet:} {Fast} \& simple resource-constrained structure learning
  of deep networks.
\newblock In {\em Proc. IEEE Conf. Computer Vision and Pattern Recognition},
  2018.

\bibitem{deepcompression}
S.~Han, H.~Mao, and W.~J. Dally.
\newblock Deep compression: {Compressing} deep neural networks with pruning,
  trained quantization and {Huffman} coding.
\newblock {\em arXiv preprint arXiv:1510.00149}, 2015.

\bibitem{NetPrune}
S.~Han, J.~Pool, J.~Tran, and W.~Dally.
\newblock Learning both weights and connections for efficient neural network.
\newblock In {\em Proc. Advances in Neural Information Processing Systems},
  pages 1135--1143, 2015.

\bibitem{resnet}
K.~He, X.~Zhang, S.~Ren, and J.~Sun.
\newblock Deep residual learning for image recognition.
\newblock In {\em Proc. IEEE Conf. Computer Vision and Pattern Recognition},
  pages 770--778, 2016.

\bibitem{amc}
Y.~He, J.~Lin, Z.~Liu, H.~Wang, L.-J. Li, and S.~Han.
\newblock {AMC}: {AutoML} for model compression and acceleration on mobile
  devices.
\newblock In {\em Proc. European Conf. Computer Vision}, pages 784--800, 2018.

\bibitem{smbo1}
J.~M. Hern{\'a}ndez-Lobato, M.~A. Gelbart, R.~P. Adams, M.~W. Hoffman, and
  Z.~Ghahramani.
\newblock A general framework for constrained {Bayesian} optimization using
  information-based search.
\newblock {\em J. Machine Learning Research}, 17(1):5549--5601, 2016.

\bibitem{mobilenet}
A.~G. Howard, M.~Zhu, B.~Chen, D.~Kalenichenko, W.~Wang, T.~Weyand,
  M.~Andreetto, and H.~Adam.
\newblock {MobileNets:} {Efficient} convolutional neural networks for mobile
  vision applications.
\newblock {\em arXiv preprint arXiv:1704.04861}, 2017.

\bibitem{condensenet}
G.~Huang, S.~Liu, L.~van~der Maaten, and K.~Q. Weinberger.
\newblock {CondenseNet}: {An} efficient {DenseNet} using learned group
  convolutions.
\newblock {\em arXiv preprint arXiv:1711.09224}, 2017.

\bibitem{binary}
I.~Hubara, M.~Courbariaux, D.~Soudry, R.~El-Yaniv, and Y.~Bengio.
\newblock Binarized neural networks.
\newblock In {\em Proc. Advances in Neural Information Processing Systems},
  pages 4107--4115, 2016.

\bibitem{squeezenet}
F.~N. Iandola, S.~Han, M.~W. Moskewicz, K.~Ashraf, W.~J. Dally, and K.~Keutzer.
\newblock {SqueezeNet}: {AlexNet-level} accuracy with 50x fewer parameters and
  $<$0.5 {MB} model size.
\newblock {\em arXiv preprint arXiv:1602.07360}, 2016.

\bibitem{progressive}
C.~Liu, B.~Zoph, J.~Shlens, W.~Hua, L.-J. Li, L.~Fei-Fei, A.~Yuille, J.~Huang,
  and K.~Murphy.
\newblock Progressive neural architecture search.
\newblock {\em arXiv preprint arXiv:1712.00559}, 2017.

\bibitem{darts}
H.~Liu, K.~Simonyan, and Y.~Yang.
\newblock Darts: {Differentiable} architecture search.
\newblock {\em arXiv preprint arXiv:1806.09055}, 2018.

\bibitem{shufflenetv2}
N.~Ma, X.~Zhang, H.-T. Zheng, and J.~Sun.
\newblock {ShuffleNet V2}: {Practical} guidelines for efficient {CNN}
  architecture design.
\newblock {\em arXiv preprint arXiv:1807.11164}, 2018.

\bibitem{hardware}
D.~Marculescu, D.~Stamoulis, and E.~Cai.
\newblock {Hardware-aware} machine learning: {Modeling} and optimization.
\newblock {\em arXiv preprint arXiv:1809.05476}, 2018.

\bibitem{pytorch}
A.~Paszke, S.~Gross, S.~Chintala, G.~Chanan, E.~Yang, Z.~DeVito, Z.~Lin,
  A.~Desmaison, L.~Antiga, and A.~Lerer.
\newblock Automatic differentiation in {PyTorch}.
\newblock In {\em Proc. Neural Information Processing Systems Workshop on
  Autodiff}, 2017.

\bibitem{mobilenetv2}
M.~Sandler, A.~Howard, M.~Zhu, A.~Zhmoginov, and L.-C. Chen.
\newblock Inverted residuals and linear bottlenecks: {Mobile} networks for
  classification, detection and segmentation.
\newblock {\em arXiv preprint arXiv:1801.04381}, 2018.

\bibitem{vgg}
K.~Simonyan and A.~Zisserman.
\newblock Very deep convolutional networks for large-scale image recognition.
\newblock {\em arXiv preprint arXiv:1409.1556}, 2014.

\bibitem{adaptiveGA}
M.~Srinivas and L.~M. Patnaik.
\newblock Adaptive probabilities of crossover and mutation in genetic
  algorithms.
\newblock {\em IEEE Trans. Systems, Man, and Cybernetics}, 24(4):656--667,
  1994.

\bibitem{bayesian_op}
D.~Stamoulis, E.~Cai, D.-C. Juan, and D.~Marculescu.
\newblock Hyperpower: Power-and memory-constrained hyper-parameter optimization
  for neural networks.
\newblock In {\em Proc. IEEE Europe Conf. \& Exihibition on Design, Automation
  \& Test}, pages 19--24, 2018.

\bibitem{mnasnet}
M.~Tan, B.~Chen, R.~Pang, V.~Vasudevan, and Q.~V. Le.
\newblock {MnasNet}: {Platform-aware} neural architecture search for mobile.
\newblock {\em arXiv preprint arXiv:1807.11626}, 2018.

\bibitem{structured_sparsity}
W.~Wen, C.~Wu, Y.~Wang, Y.~Chen, and H.~Li.
\newblock Learning structured sparsity in deep neural networks.
\newblock In {\em Proc. Advances in Neural Information Processing Systems},
  pages 2074--2082, 2016.

\bibitem{shift}
B.~Wu, A.~Wan, X.~Yue, P.~Jin, S.~Zhao, N.~Golmant, A.~Gholaminejad,
  J.~Gonzalez, and K.~Keutzer.
\newblock Shift: {A} zero {FLOP}, zero parameter alternative to spatial
  convolutions.
\newblock {\em arXiv preprint arXiv:1711.08141}, 2017.

\bibitem{shufflenet}
Z.~Xiangyu, Z.~Xinyu, L.~Mengxiao, and S.~Jian.
\newblock {ShuffleNet}: {An} extremely efficient convolutional neural network
  for mobile devices.
\newblock In {\em Proc. IEEE Conf. Computer Vision and Pattern Recognition},
  2017.

\bibitem{energy_aware}
T.-J. Yang, Y.-H. Chen, and V.~Sze.
\newblock Designing energy-efficient convolutional neural networks using
  energy-aware pruning.
\newblock {\em arXiv preprint arXiv:1611.05128}, 2016.

\bibitem{netadapt}
T.-J. Yang, A.~Howard, B.~Chen, X.~Zhang, A.~Go, M.~Sandler, V.~Sze, and
  H.~Adam.
\newblock {NetAdapt}: {Platform-aware} neural network adaptation for mobile
  applications.
\newblock In {\em Proc. European Conf. Computer Vision}, volume~41, page~46,
  2018.

\bibitem{netprune2}
T.~Zhang, K.~Zhang, S.~Ye, J.~Li, J.~Tang, W.~Wen, X.~Lin, M.~Fardad, and
  Y.~Wang.
\newblock {ADAM-ADMM}: {A} unified, systematic framework of structured weight
  pruning for {DNNs}.
\newblock {\em arXiv preprint arXiv:1807.11091}, 2018.

\bibitem{nas1}
Y.~Zhou, S.~Ebrahimi, S.~{\"O}. Ar{\i}k, H.~Yu, H.~Liu, and G.~Diamos.
\newblock Resource-efficient neural architect.
\newblock {\em arXiv preprint arXiv:1806.07912}, 2018.

\bibitem{tenary}
C.~Zhu, S.~Han, H.~Mao, and W.~J. Dally.
\newblock Trained ternary quantization.
\newblock {\em arXiv preprint arXiv:1612.01064}, 2016.

\bibitem{nasnet}
B.~Zoph, V.~Vasudevan, J.~Shlens, and Q.~V. Le.
\newblock Learning transferable architectures for scalable image recognition.
\newblock {\em arXiv preprint arXiv:1707.07012}, 2(6), 2017.

\end{thebibliography}
}

\end{document}